\documentclass[10pt,twocolumn,letterpaper]{article}

\usepackage{iccv}
\usepackage{times}
\usepackage{epsfig}
\usepackage{graphicx}
\usepackage{amsmath}
\usepackage{amssymb}
\usepackage{algorithm}
\usepackage{algpseudocode}
\usepackage{multirow}
\usepackage{booktabs}

\usepackage[pagebackref=true,breaklinks=true,letterpaper=true,colorlinks,bookmarks=false]{hyperref}

\iccvfinalcopy 

\ificcvfinal\fi

\begin{document}

\title{Multimodal Optimal Transport-based Co-Attention Transformer with Global Structure Consistency for Survival Prediction}


\author{Yingxue Xu \quad Hao Chen\thanks{Corresponding author} \\
The Hong Kong University of Science and Technology \\
{\tt\small \{yxueb, jhc\}@cse.ust.hk}
}

\maketitle
\ificcvfinal\thispagestyle{empty}\fi

\begin{abstract}
    Survival prediction is a complicated ordinal regression task that aims to predict the ranking risk of death, which generally benefits from the integration of histology and genomic data. Despite the progress in joint learning from pathology and genomics, existing methods still suffer from challenging issues: 1) Due to the large size of pathological images, it is difficult to effectively represent the gigapixel whole slide images (WSIs). 2) Interactions within tumor microenvironment (TME) in histology are essential for survival analysis. Although current approaches attempt to model these interactions via co-attention between histology and genomic data, they focus on only dense local similarity across modalities, which fails to capture global consistency between potential structures, i.e. TME-related interactions of histology and co-expression of genomic data. To address these challenges, we propose a Multimodal Optimal Transport-based Co-Attention Transformer framework with global structure consistency, in which optimal transport (OT) is applied to match patches of a WSI and genes embeddings for selecting informative patches to represent the gigapixel WSI. More importantly, OT-based co-attention provides a global awareness to effectively capture structural interactions within TME for survival prediction. To overcome high computational complexity of OT, we propose a robust and efficient implementation over micro-batch of WSI patches by approximating the original OT with unbalanced mini-batch OT. Extensive experiments show the superiority of our method on five benchmark datasets compared to the state-of-the-art methods. The code is released\footnote{\url{https://github.com/Innse/MOTCat}}.
\end{abstract}

\section{Introduction}
\begin{figure}
    \centering
    \includegraphics[scale=0.455]{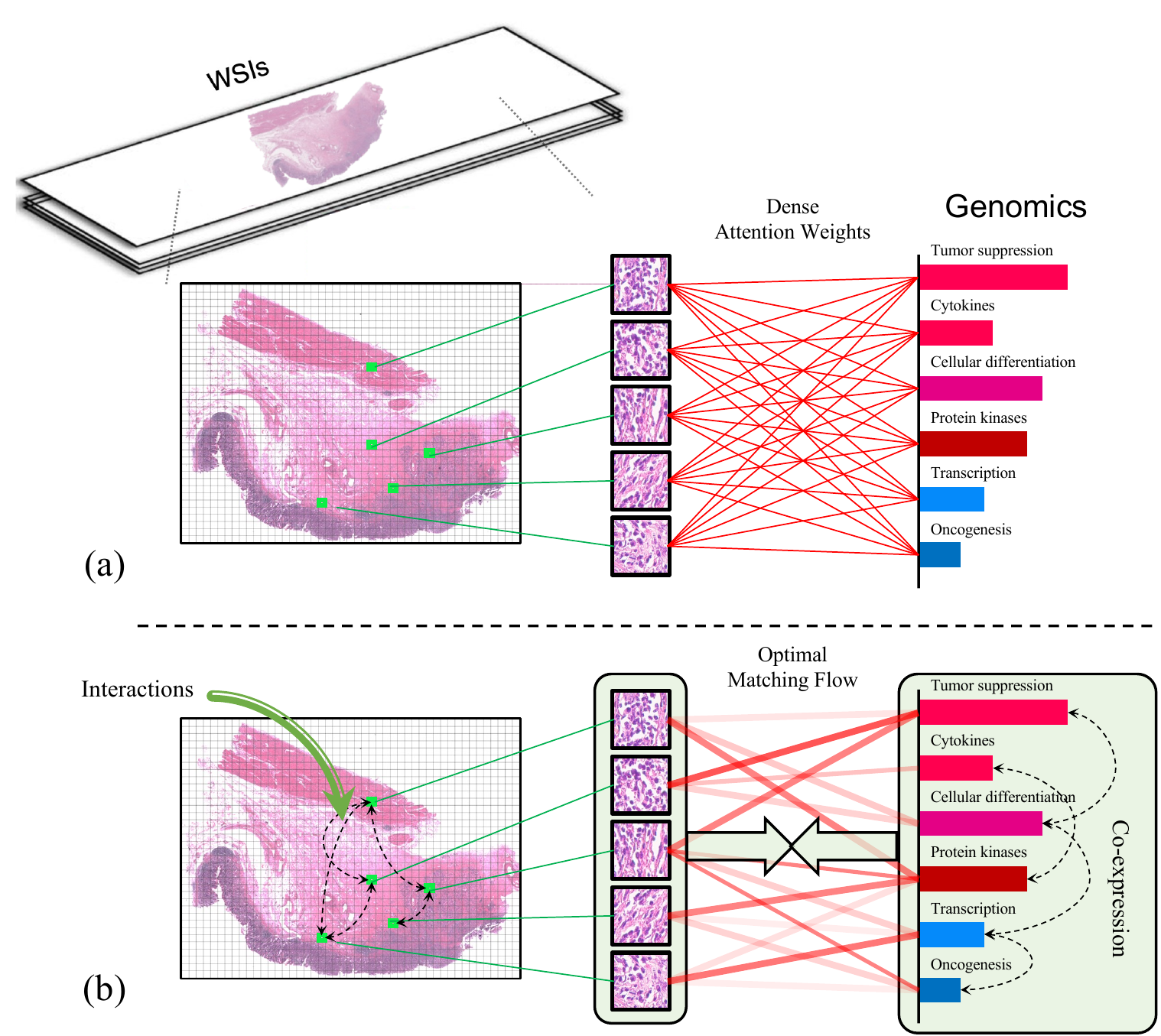}
    \caption{Comparison between (a) Co-Attention and (b) OT-based Co-Attention, where (a) learns weights of instances by dense attention, and (b) identifies informative instances by optimal transport matching flow from a global perspective, which enforces considering potential structures of each modality, i.e. interactions within WSIs and co-expression within genomics.}
    \label{fig:idea}
\end{figure}
Survival prediction is a complex ordinal regression task that aims to estimate the relative risk of death in cancer prognosis, which generally integrates the manual assessment of qualitative morphological information from pathology data and quantitative molecular profiles from genomic data in clinical settngs~\cite{chen2022pan,zuo2022identify,richard2022tmi}. Despite recent advances in multimodal learning for histology and genomics~\cite{chen2021multimodal,zuo2022identify,li2022hfbsurv,qiu2023deep}, there still exists several open issues. Among others, one daunting challenge is to capture key information from heterogeneous modalities for effective fusion. Particularly, due to the large size of about 500,000 $\times$ 500,000 pixels, it is challenging to effectively represent gigapixel whole slide images without losing key information. Furthermore, visual concepts of tumor microenvironment (TME) within pathological images are verified to have significant associations with survival analysis in various cancer types~\cite{wang2022efficient,abduljabbar2020geospatial,kuroda2021tumor}, e.g.  cellular components including fibroblast cells and various immune cells that can alter cancer cell behaviors~\cite{oya2020tumor}. However, the TME-related patches occupy only a tiny proportion of the entire WSI, leading to a fine-grained visual recognition problem~\cite{zhu2017wsisa} that is indiscernible by conventional multimodal learning.

Recently attention-based multiple instance learning (MIL)~\cite{ilse2018attention,li2021dual} has provided a typical solution to identify informative instances, in which a WSI is formulated as a bag of patches (instances) and an attention score is assigned for each instance as a weight for selection. In multimodal learning, genomic data has been applied to guide the selection of TME-related instances by co-attention mechanism across modalities~\cite{chen2021multimodal}, as genes expression might correspond to some morphological characteristics revealed in pathological TME~\cite{lin2019ghrelin,wang2019mmp}. They~\cite{chen2021multimodal} densely calculate similarity scores for each pair of pathology and genomic instances as weights of selection to capture fine-grained visual information in WSIs, as shown in Fig.~\ref{fig:idea} (a). However, this type of approaches with a local view may not thoroughly learn information about TME, since they ignore global potential structure~\cite{zuo2022identify} within modality, e.g. interactions within TME for histology and co-expression for genomics~\cite{liu2021tumor,zhou2021computational}.

Fruitful works ~\cite{liu2021tumor,oya2020tumor,chen2021multimodal} demonstrate that the interactions within TME are important indicators affecting survival outcomes, e.g. co-occurrence of tumor cells with tumor-infiltrating lymphocytes (TILs) of the immune system is a positive prognostic indicator. However, these collaborative components in TME might be spatially dispersed throughout the entire WSI, which indicates long-range structural associations in WSI, as shown in Fig.~\ref{fig:idea} (b). On the other hand, genes co-expression~\cite{zhang2014normalized,zuo2022identify} also suggests a potential structure. There might be intrinsic consistency between these two potential structures~\cite{zuo2022identify,liu2021tumor}, as some studies have verified that genomic mutations can alter normal functions and biological processes of TILs within TME~\cite{zhou2021computational,zeng2021exploration}. By leveraging the global consistency between their potential structures to match histology and genomic data, it is more likely to identify TME-related patches from WSIs. However, existing co-attention-based multimodal learning focuses on only dense local similarity, neglecting the global coherence of potential structures.

To address these challenges, we propose a \textbf{M}ultimodal \textbf{O}ptimal \textbf{T}ransport-based \textbf{C}o-\textbf{A}ttention \textbf{T}ransformer (MOTCat) framework with global structure consistency, in which optimal transport-based co-attention is applied to match instances between histology and genomics from a global perspective, as shown in Fig.~\ref{fig:idea} (b). Optimal transport (OT)~\cite{duan2022multi,caootkge2022otkge,wang2021wasserstein}, as a structural matching approach, is able to generate an optimal matching solution with the overall minimum matching cost, based on local pairwise cost between instances of histology and genomics. As a result, instances of a WSI that have high global structure consistency with genes co-expressions can be identified to represent the WSI. These instances might be more strongly associated with TME that contributes to survival prediction. In this way, the aforementioned two challenges can be addressed by OT-based Co-Attention simultaneously.

Compared with the conventional co-attention mechanism, the advantages of the proposed OT-based co-attention are three-fold. First, optimal transport provides the instances matching process with global awareness. Marginal constraints of total mass equality enforce a trade-off of instances within modality for transporting during optimization, instead of only considering the local similarity of pairwise instances as conventional co-attention does. It enables the modeling of structural interactions of WSIs and co-expressions of genomics, which is beneficial to survival prediction. Second, the learned patch-level attention score is not a rigorous metric for selecting informative instances under weak supervision~\cite{zhang2022dtfd}, i.e. slide-level survival time, especially with only a small number of WSI samples. As an alternative, the optimal matching flow is the rigorous closed-form solution to OT problem without accessing any label, achieved by solving the Linear Programming problem. Last, the optimal matching flow provides a transformation between two modalities under preserving the potential structure, which reduces cross-modal heterogeneity gap.

Nevertheless, due to a massive number of patches from gigapixel WSIs, OT-based co-attention might suffer from high computational complexity, preventing from applying OT to pathology data in practice. To address this, we propose a robust and efficient implementation of OT-based co-attention for matching histology and genomics. Specifically, we split all instances of a WSI into a subset termed Micro-Batch and get the averaged result as a proxy to approximate the solution to the original OT problem over all instances by unbalanced optimal transport ~\cite{fatras2021unbalanced}, which can provide a more robust approximation to the subset sampling with the theoretical and experimental guarantee.

It is worth noting that the proposed method is a generalized multimodal learning framework that captures potential structure across modalities with global structure consistency. The contributions of this work are summarized as follows: (1) we propose a novel multimodal OT-based Co-Attention Transformer with global structure consistency, where OT-based co-attention is used to identify informative instances by global optimal matching, which allows modeling interactions of histology and co-expression of genomics. (2) We propose a robust and efficient implementation of OT-based co-attention over Micro-Batch. (3) Extensive experiments on five benchmark datasets show significant improvement over state-of-the-art methods. 

\section{Related Work}
\subsection{Multimodal Learning in Healthcare}
In clinical settings, the patient state is often characterized by a spectrum of various modalities~\cite{lipkova2022artificial}, such as pathology~\cite{zhu2017wsisa,lu2021data}, radiology~\cite{wang2020preoperative,zheng2022multi}, genomics~\cite{liberzon2015molecular,zeng2021exploration,liu2021tumor}, etc. Multimodal learning has shown superiority by leveraging the complementary information from multimodal data~\cite{lipkova2022artificial}, in which the key issue is to overcome data heterogeneity for better feature fusion. Most existing methods can be roughly classified into three categories: early fusion~\cite{huang2020fusion,wang2021gpdbn}, late fusion~\cite{joo2021multimodal,richard2022tmi}, and intermediate fusion~\cite{chen2021multimodal,kumar2019co}. Early fusion is to aggregate all modalities at the input level. The most straightforward solution is fusion operators such as concatenation~\cite{huang2020fusion}, Kronecker Product~\cite{ramachandram2017deep}, etc. However, it neglects intra-modality dynamics~\cite{lipkova2022artificial}. On the contrary, late fusion~\cite{gadzicki2020early} integrates the predictions from individually separated models at the decision level for the final decision, which cannot fully explore cross-modal interactions. Recently, intermediate fusion has attracted much interest by capturing cross-modal interconnections at different levels, where the typical one is attention-based multimodal fusion~\cite{chen2021multimodal}. For example, Zhu et al. ~\cite{zhu2022multimodal} proposed a triplet attention to integrate MRI and diffusion tensor imaging (DTI) for epilepsy diagnosis. IMGFN~\cite{zuo2022identify} designed a graph attention network for survival prediction. MCAT~\cite{chen2021multimodal} proposed a co-attention to identify informative instances of gigapixel WSI with genomic features as queries. HFBSurv~\cite{li2022hfbsurv} progressively integrated multimodal information from the low level to the high level for survival analysis. Note that our method MOTCat belongs to one of the intermediate fusion approaches, which explores the global consistency of potential structure by optimal transport.
\subsection{Multiple Instance Learning in Pathology}
Since it is difficult to represent WSIs of large size, MIL algorithms have achieved remarkable progress in pathology, in which WSIs are often formulated as a bag of pathology patches. MIL in pathology generally can be divided into two categories: instance-based algorithms ~\cite{campanella2019clinical,xu2019camel,chikontwe2020multiple} and embedding-based algorithms~\cite{ilse2018attention,hashimoto2020multi,lu2021data}. The former aims to select a small number of instances within each WSI for training an aggregation model. The latter one is to map each instance into a low-dimension fixed-length embedding offline, and then to learn a bag-level representation based on instance-level embeddings. There are several strategies for aggregating instance-level embeddings. One is clustering-based methods~\cite{sharma2021cluster,wang2019weakly}, where all the instance-level embeddings are clustered to several centroids that are integrated for the final prediction. The most common strategy is attention-based MIL (AB-MIL)~\cite{ilse2018attention,li2021dual} that assigns a weight for each instance, in which various approaches are developed differing in the ways to generate the weighting values. For example, one of the first AB-MIL~\cite{ilse2018attention} used a side-branch network for generating attention weights, followed by DS-MIL~\cite{li2021dual} adopted cosine distance to measure the similarity of instances as weights. Recently, transformer-based MIL like TransMIL~\cite{shao2021transmil} has leveraged the self-attention mechanism to explore long-range interactions in WSI. Furthermore, DTFD-MIL~\cite{zhang2022dtfd} proposed the double-tier MIL framework for more rigorous instance weights. Unlike AB-MIL, we use optimal match flow generated by OT between pathological and genomic instances to identify instances of WSI that have global potential structure consistency with that of genomics.
\subsection{Survival Prediction}
Survival outcome prediction, also known as time-to-event analysis~\cite{2011Survival}, aims to predict the probability of experiencing an event of interest, e.g. death event in the clinical setting, before a specific time under both right-censored and uncensored data. Right-censored data refers to cases where the outcome event cannot be observed within the study period~\cite{2011Survival}. Before the deep learning era, statistical models dominated survival analysis. The typical one is Cox Proportional Hazards (CoxPH) model~\cite{david1972regression} that characterizes the hazard function, the risk of death, by an exponential linear model. DeepSurv~\cite{faraggi1995neural} is the first work incorporating deep learning into survival analysis, which models the risk function by several fully connected layers and feeds it to the CoxPH model for the final hazard function. Furthermore, DeepConvSurv~\cite{zhu2016deep} first attempted to use pathological images for a deep survival model. Recent works~\cite{li2022hfbsurv,chen2022pan,qiu2023deep,richard2022tmi,chen2021multimodal} tend to directly estimate the hazard function by deep learning models without any statistical assumption. Histology and genomics data are often integrated as the gold standard for predicting survival outcomes with promising performance~\cite{richard2022tmi,chen2022pan,zuo2022identify}. Similarly, our method incorporates these two modalities for better survival prediction by learning their global structural coherence.

\section{Method}
\subsection{Overview and Problem Formulation}
\begin{figure*}[t]
    \centering
    \includegraphics[scale=0.52]{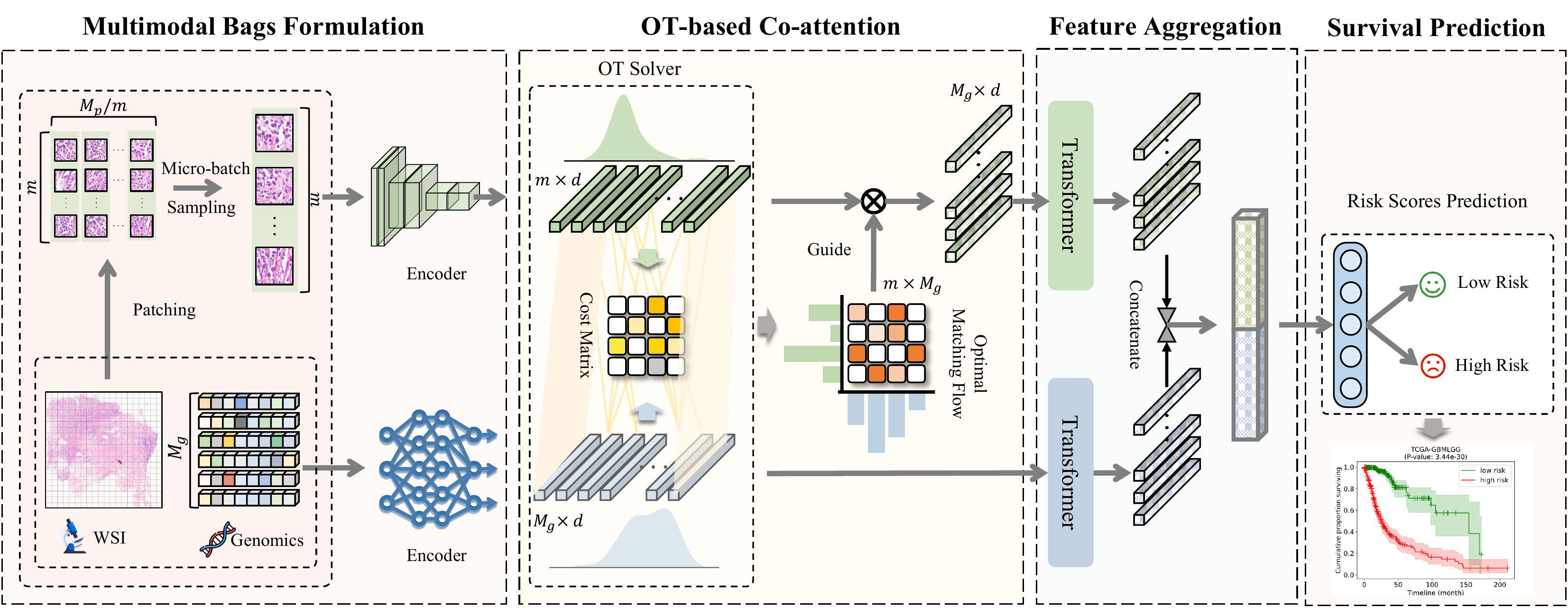}
    \caption{Overview of Multimodal Optimal Transport-based Co-Attention Transformer (MOTCat) architecture, in which both modalities are formulated as bags representations and a micro-batch of the WSI bag is sampled as pathological input. Then OT-based Co-attention is to calculate optimal matching flow for identifying informative instances with global structure consistency. After aggregating selected instances, features from both modalities are concatenated for survival prediction.}
    \label{fig:overview}
\end{figure*}
The overall framework of the proposed method is shown in Fig.~\ref{fig:overview}. In this work, we aim to estimate the hazard functions $h_n(t)$ considered as the risk probability of experiencing death event in a small interval after time point $t$ for $n$-th patient, based on pathology data $\mathbf{X}_n^p$ and genomic data $\mathbf{X}_n^g$.

For pathology, we formulate each WSI as a "bag" structure in a conventional multiple instance learning (MIL) setting, where a bag consists of a collection of instances that are patched from the tissue-containing image regions of a WSI, and instance-level features are extracted from patch images. Genomic data is also organized as a "bag" based on the biological functional impact of genes. For each instance of the genomic bag, we adopt an encoder to extract genes embeddings. The bags formulations for histology and genomics are illustrated in Sec.~\ref{sec: formulate wsi} and ~\ref{sec: formulate genes}, respectively.

Then, Optimal Transport-based Co-Attention is applied to match instances of a WSI and the corresponding genomic instances to model pathological interactions and genomic co-expressions, so that the instances with high structure consistency can be identified to represent the whole modality. This part will be demonstrated in Sec.~\ref{sec:MOTCat}.

After that, a Transformer encoder will integrate all instances after selection for each modality into a bag-level representation, in which each instance is treated as a token. The bag-level representation will be concatenated into the multimodal features for predicting a hazard function to get the ordinal survival risk. The details will be described in Sec.~\ref{sec:survival prediction}. Finally, optimization process over Micro-Batch and the corresponding computational complexity analysis will be presented in Sec.~\ref{sec:opt}.

\subsubsection{WSI Bags Formulation}
\label{sec: formulate wsi}
We formulate the learning process of pathology images (WSIs) as a weakly-supervised MIL task, in which each WSI $\mathbf{X}_n^p$ is represented as a bag with access to only bag-level (slide-level) labels. Given a bag $\mathbf{X}_n^p = \{ \mathbf{x}_{n,i}^p \}_{i=1}^{M_p}$ of $M_p$ permutation-invariant instances, the goal is to encode instance-level features of a WSI into a compact bag-level feature embedding and then assign a bag-level label to it. To this end, we apply a CNN encoder $f_p(\cdot)$ for each patch of the WSI to obtain a bag $\mathbf{B}_n^p$ of instance-level features:
\begin{equation}
\label{equ:bag_p}
    \mathbf{B}_n^p = \{f_{p}(\mathbf{x}_{n,i}^p): \mathbf{x}_{n,i}^p \in \mathbf{X}_n^p\}=\{\mathbf{b}^p_{n,i}\}_{i=1}^{M_p},
\end{equation}
where $\mathbf{B}_n^p \in \mathbb{R}^{M_p \times d}$ contains $d$-dimensional instances and $\mathbf{b}^p_{n,i}=f_{p}(\mathbf{x}_{n,i}^p)$. After identifying informative instances of $\mathbf{B}_n^p$ by OT-based Co-Attention, we construct a new bag $\hat{\mathbf{B}}_n^p \in \mathbb{R}^{\hat{M}_p \times d} $ of features with $\hat{M}_p$ instances,  which will be illustrated in detail in Section ~\ref{sec:MOTCat}. Then, a transformer encoder is used as a global aggregation model $\mathcal{T}_p(\cdot)$ to obtain the bag-level representation $H_n^p = \mathcal{T}_p(\hat{\mathbf{B}}_n^p)$ for the subsequent features fusion.

\begin{algorithm}[ht]
\caption{The MOTCat Algorithm}
\label{alg:opt}
\begin{algorithmic}[1]
\Require A WSI bag as $\mathbf{X}_n^p=\{ \mathbf{x}_{n,i}^p \}_{i=1}^{M_p}$; A genomic bag as $\mathbf{X}_n^g=\{ \mathbf{x}_{n,j}^g \}_{j=1}^{M_g}$. $m$ refers to the size for a micro-batch of a WSI.
\Ensure The optimal parameters of model; \\

// Sample micro-batch from the WSI bag $X_n^{p,m}=\{ \mathbf{x}_{n,i}^p \}_{i=1}^{m}$ each time.
\For{$X_n^{p,m}$ in $\mathbf{X}_n^p$}
\State Fix model parameters $f_p$, $f_g$, $\mathcal{T}_p$ and $\mathcal{T}_g$.
\State // To extract a bag of instance-level features
\State // Eq.~\ref{equ:bag_p} over micro-batch of histology
\State $B_n^{p,m} = \{f_p(\mathbf{x}_{n,i}^p):\mathbf{x}_{n,i}^p \in X_n^{p,m} \}$
\State // Eq.~\ref{equ:bag_g} of genomics
\State $\mathbf{B}_n^g=\{f_g^j(\mathbf{x}_{n,j}^g): \mathbf{x}_{n,j}^g\in \mathbf{X}_n^g \}$
\State Optimize $\mathbf{P}_n^m$ via Sinkhorn-Knopp matrix scaling algorithm~\cite{chizat2018scaling,frogner2015learning}:
\State \ \ \ \ $\mathbf{P}_n^m \gets \arg\min_{\mathbf{P}_n^m} \mathcal{W}^m(B_n^{p,m}, \mathbf{B}_n^{g})$ // Eq.~\ref{equ:ot_mb}
\State Fix $\mathbf{P}_n^m$ and select instances from $B_n^{p,m}$ by:
\State \ \ \ \ $\hat{B}_n^{p,m}={\mathbf{P}_n^m}^\top B_n^{p,m}$
\State // To get bag-level representation
\State $H_n^{p,m}=\mathcal{T}_p\left(\hat{B}_n^{p,m}\right)$ // for histology
\State $H_n^g=\mathcal{T}_g\left(\hat{\mathbf{B}}_n^g\right)$ // for genomics
\State $H_n^{m} = H_n^{p,m}\bowtie H_n^g$ // Concatenation
\State Calculate the overall loss function $\mathcal{L}$ of Eq.~\ref{equ:loss}.
\State Update parameters of $f_g$, $\mathcal{T}_p$, $\mathcal{T}_g$ and the last fully connected layers of $f_p$ by Adam.
\EndFor
\end{algorithmic}
\end{algorithm}

\subsubsection{Genomic Bags Formulation}
\label{sec: formulate genes}
Genomic data consists of $1\times 1$ attributes, such as mutation status, transcript abundance (RNA-Seq), copy number variation (CNV) and other molecular characterizations. Following the organization of genomic data in the previous work~\cite{chen2021multimodal,liberzon2015molecular} that categories them according to the biological functional impact, the expressive genomic bag can be organized as $\mathbf{X}_n^g=\{ \mathbf{x}_{n,j}^g\}_{j=1}^{M_g} \in \mathbb{R}^{M_g \times d_j}$ with $M_g$ instances of unique functional categories. For $j$-th category, the corresponding genomic embedding consists of $d_j$-dimensional attributes encoded by a separated network $f_g^j(\cdot)$. Similarly, a bag $\mathbf{B}_n^g\in \mathbb{R}^{M_g \times d}$ of genomic data with $M_g$ instances can be constructed as follows:
\begin{equation}
\label{equ:bag_g}
    \mathbf{B}_n^g=\{f_g^j(\mathbf{x}_{n,j}^g): \mathbf{x}_{n,j}^g\in \mathbf{X}_n^g \}=\{\mathbf{b}^g_{n,j}\}_{j=1}^{M_g},
\end{equation}
which is aggregated to the bag-level embedding by a transformer encoder $H_n^g=\mathcal{T}_g(\mathbf{B}_n^g)$ as well.

\subsubsection{Multimodal Survival Prediction Formulation}
\label{sec:survival prediction}
Survival outcome prediction aims to predict the risk probability of an outcome event occurring before a specific time, in which the outcome event is not always observed, leading to a right-censored event. In this task, let $c\in \{0,1\}$ be censor status indicating whether the outcome event is right-censored ($c=1$) or not ($c=0$), and $t\in \mathbb{R}^{+}$ refers to the overall survival time (in months).
Given a training set of $N$ pathology-genomics pairs $\mathcal{D}=\{(\mathbf{X}_n^p, \mathbf{X}_n^g),c_n,t^{n} \}_{n=1}^{N}$, we can acquire the bag-level features $H^p_n$ of WSIs and $H^g_n$ of genomic data for $n$-th patient data $\mathbf{X}_n=((\mathbf{X}_n^p, \mathbf{X}_n^g),c_n,t^{n})$, as described in Section ~\ref{sec: formulate wsi} and ~\ref{sec: formulate genes}. After integrating $H^p_n$ and $H^g_n$ into the multimodal features $H_n$ by concatenation, we estimate the hazard function $h_n(t|H_n)=h_n(T=t|T\geq t, H_n)\in [0,1]$ which is viewed as the probability of death event occurring at around time point $t$. Instead of predicting the overall survival time $t^n$, the survival prediction task is to estimate the ordinal risk of death via the cumulative survival function:
\begin{equation}
    S_n(t|H_n) = \prod_{z=1}^{t}(1-h_n(z|H_n))
\end{equation}
\subsection{Optimal Transport-based Co-Attention}
\label{sec:MOTCat}
To identify TME-related instances of WSIs, we utilize the global potential structure consistency between TME interactions of pathology and genes co-expressions as the evidence of selecting instances, in which optimal transport is used to learn the optimal matching flow between a WSI feature bag $\mathbf{B}_n^p \in \mathbb{R}^{M_p \times d}$ and a genomic bag $\mathbf{B}_n^g \in \mathbb{R}^{M_g \times d}$. Formally, an optimal transport from the WSI feature bag $\mathbf{B}_n^p=[\mathbf{b}^p_{n,1},\mathbf{b}^p_{n,2},\ldots,\mathbf{b}^p_{n,M_p}]$ to the genomic embedding bag $\mathbf{B}_n^g=[\mathbf{b}^g_{n,1},\mathbf{b}^g_{n,2},\ldots,\mathbf{b}^g_{n,M_g}]$ can be defined by the discrete Kantorovich formulation~\cite{kantorovich2006translocation} to search the overall optimal matching flow $\mathbf{P}_n$ between $\mathbf{B}_n^p$ and $\mathbf{B}_n^g$:
\begin{equation}
    \mathcal{W}(\mathbf{B}_n^p, \mathbf{B}_n^g)=\min_{\mathbf{P}_n\in \Pi(\mu_p, \mu_g)} <\mathbf{P}_n,\mathbf{C}_n>_F
\label{equ:ot}
\end{equation}
where $\mathbf{C}_n\geq 0 \in \mathbb{R}^{M_p \times M_g}$ is a cost matrix given by $\mathbf{C}_n^{u,v}=c(\mathbf{b}^p_{n,u},\mathbf{b}^g_{n,v})$ with a ground distance metric $c(\cdot)$, such as $l_2$-distance, that measures the distance of local pairwise instances $\mathbf{b}^p_{n,u} \in \mathbf{B}_n^p$ in the WSI bag and the genomic one of a unique functional category $\mathbf{b}^g_{n,v}\in \mathbf{B}_n^g$. 

Here $\Pi(\mu_p,\mu_g)=\{ \mathbf{P}_n \in \mathbb{R}_{+}^{M_p \times M_g} | \mathbf{P}_n \mathbf{1}_{M_g}=\mu_p, \mathbf{P}_n^\top\mathbf{1}_{M_p} = \mu_g \}$ involves the marginal constraints of total mass equality between marginal distributions, $\mu_p$ and $\mu_g$, for the WSI bag and the genomic bag, and $\mathbf{1}_k$ is a $k$-dimensional vector of ones. Specifically, $\Pi(\mu_p,\mu_g)$ refers to the set of joint probabilistic couplings between the two marginal empirical distributions of pathology data and genomic data. Intuitively, it describes how to distribute instances of a WSI bag $\mathbf{B}_n^p$ to the genomic embedding of a genomic bag $\mathbf{B}_n^g$ based on the cost matrix $\mathbf{C}_n$, under the marginal constraints of total mass equality between distributions $\mu_p$ and $\mu_g$. Note that $<\cdot>_F$ refers to the Frobenius dot product, and thus Eq. ~\ref{equ:ot} encourages to achieve the overall minimum matching cost by finding optimal matching flow based on local pairwise similarity. In this way, the optimal transport of Eq.~\ref{equ:ot} is able to enforce a trade-off of instances within modality, which allows the model to capture the potential structure of visual interactions for histology and co-expressions for genomic data.

Once acquiring the optimal matching flow $\mathbf{P}_n^*$, informative instances of a WSI are identified by $\hat{\mathbf{B}}_n^p=\mathbf{P}_n^\top \mathbf{B}_n^p$ to represent the WSI, which also aligns pathology distribution to genomics distribution under preserving the potential structure across modalities for alleviating heterogeneity.

As such, the proposed OT-based Co-Attention simultaneously addresses two issues: not only does it achieve the effective representation for gigapixel WSIs by identifying informative instances via OT matching, but marginal constraints of OT enable the selected instances to have global structure consistency with genomic data.

\subsection{Optimization over Micro-Batch}
\label{sec:opt}
Due to the large size of the WSI bag, it is difficult to apply optimal transport for matching histology data and genomic data. Recent work~\cite{zhang2022dtfd} has validated that MIL on WSI can benefit from multiple subsets of a bag randomly sampled from the original WSI bag. It provides a way to train the model over Micro-Batch of a WSI, where micro-batch is defined as a subset sampled from a bag of WSI instances. Furthermore, inspired by a variant ~\cite{fatras2021unbalanced} of OT that offers a solution to approximate the original OT over mini-batches with a theoretical guarantee of convergence, we propose to use the variant UMBOT formulation~\cite{fatras2021unbalanced} over micro-batch of WSI instead of the original OT over all instances of a WSI. Then the Eq.~\ref{equ:ot} becomes:
\begin{small}
\begin{equation}
\label{equ:ot_mb}
\begin{aligned}
    &\mathcal{W}^m(B_n^{p,m}, \mathbf{B}_n^{g}) = \min_{\mathbf{P}_n^m \in \Pi(\mu_p^m, \mu_g)} <\mathbf{P}_n^m,\mathbf{C}_n^m>_F \\
    &+ \epsilon KL(\mathbf{P}_n^m|\mu_p^m\otimes\mu_g)+\tau\left( D_\phi(\mathbf{P}_{n,p}^{m}||\mu_p^m)+D_\phi(\mathbf{P}_{n,g}^{m}||\mu_g) \right)
\end{aligned}
\end{equation}
\end{small}
where $B_n^{p,m}$ is micro-batch of size $m$ sampled from $\mathbf{B}_n^{p}$. Similarly, $\mathbf{C}_n^m \in \mathbb{R}^{m\times M_g}$ is the cost matrix over micro-batch of WSI $B_n^{p,m}$ and $\mathbf{B}_n^{g}$, and the optimal matching flow $\mathbf{P}_n^m\in \mathbb{R}^{m\times M_g}$ is optimized over micro-batch as well. Here $\mathbf{P}_{n,p}^{m}$ and $\mathbf{P}_{n,g}^{m}$ represent the marginals of $\mathbf{P}_n^m$, and $D_\phi$ is Csiszàr divergences, $\tau$ and $\epsilon \geq 0$ are the coefficients of marginal penalization and entropic regularization, respectively. $KL$ refers to Kullback-Leible divergence. 

\noindent
\textbf{Computational complexity.} With the benefit of UMBOT for approximating the original OT, the computational complexity of optimization is reduced from $\mathcal{O}(M^3\log(M))$ to $\mathcal{O}(\frac{M}{m}\times m^2)=\mathcal{O}(M\times m)$, where $M=\max(M_p, M_g)$ and $m \ll M$. The whole optimization procedure of the proposed MOTCat is presented in Alg.~\ref{alg:opt}

\noindent
\textbf{Loss function.} Following the previous work~\cite{chen2021multimodal}, the overall loss function is formulated by NLL-loss~\cite{zadeh2020bias}:
\begin{equation}
\label{equ:loss}
\begin{aligned}
    \mathcal{L} = &-\frac{m}{M_p}\sum_{n=1}^{N}\sum_{X_n^{p,m}\in \mathbf{X}_n^p} c_n\cdot \log\left(S_n^m(t_n|H_n^m)\right)\\
    &-\frac{m}{M_p}\sum_{n=1}^{N}\sum_{X_n^{p,m}\in \mathbf{X}_n^p} \{ (1-c_n)\cdot \log\left(S_n^m(t_n-1|H_n^m)\right)\\
    &+(1-c_n)\cdot \log (h_n^m(t_n|H_n^m))\}
\end{aligned}
\end{equation}
where $H_n^m$ refers to the multimodal features over micro-batch formulated in line 16 of Alg.~\ref{alg:opt}. $h_n^m$ and $S_n^m$ are the hazard function and cumulative survival function over micro-batch, respectively.

\section{Experiment}
In this section, we first present datasets and settings following previous experimental protocols~\cite{chen2021multimodal,chen2022pan} for fair comparisons. Next, we demonstrate the performance results of the proposed method compared with the state-of-the-art (SOTA) methods, including unimodal and multimodal approaches. After that, we investigate the effectiveness of each component in our method and the effect on the size of micro-batch strategy. Finally, from a statistical point of view, we present Kaplan-Meier survival curves and Logrank test to show the performance of survival analysis. Interpretable visualization for histology data and genomic data are presented in supplementary materials.
\subsection{Datasets and Settings}
\begin{table*}[ht]
    \centering
    \small
    \caption{C-Index (mean $\pm$ std) performance over five cancer datasets. Patho. and Geno. refer to pathology modality and genomic modality, respectively. The best results and the second-best results are highlighted in \textbf{bold} and in \underline{underline}, respectively.}
    \begin{tabular}{l|cc|ccccc}
    \toprule
        \multirow{2}{*}{Model} & \multirow{2}{*}{Patho.} & \multirow{2}{*}{Geno.} & BLCA & BRCA & UCEC & GBMLGG & LUAD  \\
        ~&~&~& ($N=373$) & ($N=956$) & ($N=480$) & ($N=569$) & ($N=453$)\\
        \midrule
         SNN*~\cite{klambauer2017self}&~& $\checkmark$ & 0.618 $\pm$ 0.022 & 0.624 $\pm$ 0.060 & \underline{0.679 $\pm$ 0.040} & 0.834 $\pm$ 0.012 & 0.611 $\pm$ 0.047 \\
         SNNTrans*~\cite{klambauer2017self,shao2021transmil}&~& $\checkmark$ & 0.659 $\pm$ 0.032 & 0.647 $\pm$ 0.063 & 0.656 $\pm$ 0.038 & 0.839 $\pm$ 0.014 & 0.638 $\pm$ 0.022 \\
         \midrule
         AttnMIL*~\cite{ilse2018attention} & $\checkmark$ &~& 0.599 $\pm$ 0.048 & 0.609 $\pm$ 0.065 & 0.658 $\pm$ 0.036 & 0.818 $\pm$ 0.025 & 0.620 $\pm$ 0.061 \\
         DeepAttnMISL~\cite{yao2020whole} & $\checkmark$ & ~ & 0.504 $\pm$ 0.042 & 0.524 $\pm$ 0.043 & 0.597 $\pm$ 0.059 & 0.734 $\pm$ 0.029  & 0.548 $\pm$ 0.050  \\
         
         CLAM-SB*~\cite{lu2021data} & $\checkmark$ & ~ & 0.559 $\pm$ 0.034 &  0.573 $\pm$ 0.044 &  0.644 $\pm$ 0.061  &  0.779 $\pm$ 0.031 & 0.594 $\pm$ 0.063 \\
         CLAM-MB*~\cite{lu2021data} & $\checkmark$ & ~ & 0.565 $\pm$ 0.027 &  0.578 $\pm$ 0.032 &  0.609 $\pm$ 0.082  &  0.776 $\pm$ 0.034 & 0.582 $\pm$ 0.072 \\
         TransMIL*~\cite{shao2021transmil} & $\checkmark$ & ~ & 0.575 $\pm$ 0.034 & \underline{0.666 $\pm$ 0.029} & 0.655 $\pm$ 0.046 & 0.798 $\pm$ 0.043 & 0.642 $\pm$ 0.046 \\
         DTFD-MIL*~\cite{zhang2022dtfd} & $\checkmark$ & ~ & 0.546 $\pm$ 0.021 & 0.609 $\pm$ 0.059 & 0.656 $\pm$ 0.045 & 0.792 $\pm$ 0.023 & 0.585 $\pm$ 0.066 \\
         \midrule
         
         Pathomic~\cite{richard2022tmi} & $\checkmark$ & $\checkmark$ & 0.586 $\pm$ 0.062 &  -  &  -  &  0.826 $\pm$ 0.009 & 0.543 $\pm$ 0.065 \\
         
         PONET~\cite{qiu2023deep} & $\checkmark$ & $\checkmark$ & 0.643 $\pm$ 0.037 & - & - & - & 0.646 $\pm$ 0.047 \\
         Porpoise*~\cite{chen2022pan} & $\checkmark$ & $\checkmark$ & 0.636 $\pm$ 0.024 &  0.652 $\pm$ 0.042 &  \textbf{0.695 $\pm$ 0.032}  &  0.834 $\pm$ 0.017 & 0.647 $\pm$ 0.031 \\
         
         MCAT*~\cite{chen2021multimodal} & $\checkmark$ & $\checkmark$ & \underline{0.672 $\pm$ 0.032} &  0.659 $\pm$ 0.031  &  0.649 $\pm$ 0.043  &  \underline{0.835 $\pm$ 0.024} & \underline{0.659 $\pm$ 0.027} \\
         \textbf{MOTCat (Ours)} & $\checkmark$ & $\checkmark$ & \textbf{0.683 $\pm$ 0.026} &  \textbf{0.673 $\pm$ 0.006}  &  0.675 $\pm$ 0.040  &  \textbf{0.849 $\pm$ 0.028} & \textbf{0.670 $\pm$ 0.038} \\
         
         \bottomrule
    \end{tabular}
    \label{tab:sota}
\end{table*}

\noindent
\textbf{Datasets.} To demonstrate the performance of the proposed method, we conducted various experiments over five public cancer datasets from The Cancer Genome Atlas (TCGA) that contains paired diagnostic WSIs and genomic data with ground-truth survival outcome: Bladder Urothelial Carcinoma (BLCA), Breast Invasive Carcinoma (BRCA), Uterine Corpus Endometrial Carcinoma (UCEC), Glioblastoma \& Lower Grade Glioma (GBMLGG), and Lung Adenocarcinoma (LUAD). The number of cases for each cancer type is shown by $N$ in Tab.~\ref{tab:sota}. Note that cases of these cancer datasets used for all compared methods are not less than cases used in the proposed method. For genomic data, the number of unique functional categories $M_g$ is set as 6 following ~\cite{liberzon2015molecular,chen2021multimodal}, including 1) Tumor Supression, 2) Oncogenesis, 3) Protein Kinases, 4) Cellular Differentiation, 5) Transcription, and 6) Cytokines and Growth.

\noindent
\textbf{Evaluation.} For each cancer dataset, we perform 5-fold cross-validation with a 4:1 ratio of training-validation sets and report the cross-validated concordance index (C-Index) and its standard deviation (std) to quantify the performance of correctly ranking the predicted patient risk scores with respect to overall survival.

\noindent
\textbf{Implementation.} For each WSI, we first apply the OTSU's threshold method to segment tissue regions, and then non-overlapping $256\times 256$ patches are extracted from the tissue region over $20\times$ magnification. Then, we use ImageNet-pretrained~\cite{deng2009imagenet} ResNet-50~\cite{he2016identity} and a 256-d fully-connected layer as the feature encoder $f_p$ to extract the 1024-d embedding for each patch, where parameters of ResNet-50 are frozen. For genomic data, we adopt SNN~\cite{klambauer2017self} as the feature encoder $f_g$ following the setting of ~\cite{chen2021multimodal}.

During training, we follow the setting of ~\cite{chen2021multimodal} for a fair comparison. Specifically, we adopt Adam optimizer with the initial learning rate of $2\times 10^{-4}$ and weight decay of $1\times 10^{-5}$. Due to the large size of WSIs, the batch size for WSIs is 1 with 32 gradient accumulation steps, and all experiments are trained for 20 epoches. The size of Micro-Batch $m$ is set as 256. Regarding the hyper-parameters of OT in Eq.~\ref{equ:ot_mb}, the coefficient of marginal penalization $\tau$ is 0.5 for all cancer datasets, and the coefficient of entropic regularization $\epsilon$ is 0.05 for BLCA and LUAD, as well as 0.1 for BRCA, UCEC and GBMLGG.
\subsection{Results}
We compare our method against the unimodal baselines and the multimodal SOTA methods as follows:

\noindent
\textbf{Unimodal Baseline.} For genomic data, we adopt \textbf{SNN}~\cite{klambauer2017self} that has been used previously for survival outcome prediction in the TCGA~\cite{richard2022tmi,chen2021multimodal}, and \textbf{SNNTrans}~\cite{klambauer2017self,shao2021transmil} which incorporates SNN as the feature extractor and TransMIL~\cite{shao2021transmil} as the global aggregation model for MIL. For histology, we campare the SOTA MIL methods \textbf{AttnMIL}~\cite{ilse2018attention}, \textbf{DeepAttnMISL}~\cite{yao2020whole}, \textbf{CLAM}~\cite{lu2021data}, \textbf{TransMIL}~\cite{shao2021transmil}, and \textbf{DTFD-MIL}~\cite{zhang2022dtfd}.

\noindent
\textbf{Multimodal SOTA.} We compare four SOTA methods for multimodal survival outcome prediction including \textbf{MCAT}~\cite{chen2021multimodal}, \textbf{Pathomic}~\cite{richard2022tmi}, \textbf{Porpoise}~\cite{chen2022pan} and \textbf{PONET}~\cite{qiu2023deep}.

Results are shown in Tab.~\ref{tab:sota}, where the methods marked * are re-implemented and the best results are reported, and other results are obtained from their papers. Note that due to varying cancer datasets used in their works, not all results of the five datasets used in this work can be reported, and hence we use '-' to indicate the results are not reported in their papers.

\noindent
\textbf{Unimodal v.s. Multimodal.} Compared with all unimodal approaches, the proposed method achieves the highest performance in 4 out of 5 cancer datasets, indicating the effective fusion of multimodal data in our method. Note that the overall performance of genomic data is better than that of histology, which validates the reasonability of using genomic data for guiding the selection of instances in WSIs. In particular, most multimodal methods are inferior to the unimodal model of genomics in UCEC dataset, suggesting the serious challenge in multimodal fusion. Nevertheless, the proposed method can achieve the comparative performance with genomic model on UCEC.

\noindent
\textbf{Multimodal SOTA v.s. MOTCat.} In one-versus-all comparisons of multimodal models, the proposed method achieves superior performance on all benchmarks with 1.0\%-2.6\% performance gains except UCEC. In comparing the proposed method with the most similar work MCAT in multimodal fusion, our method gets better results on all datasets, indicating the effectiveness of learning the intra-modal potential structure consistency from a global perspective between TME-related interactions of histology and genomic co-expression.

\subsection{Ablation Study}
\begin{table*}[t]
    \centering
    \caption{Ablation study assessing C-Index (mean $\pm$ std) performance of MOTCat over five datasets, in which version \textbf{(c)} in bold is the proposed method with full components, and (a) and (b) refer to its variants. The best results are marked in \textbf{bold}.}
    \begin{tabular}{l|cc|ccccc}
    \toprule
        Variants & OT & MB & BLCA & BRCA & UCEC & GBMLGG & LUAD \\
        \midrule
         (a) MCAT &~& ~ & 0.672 $\pm$ 0.032 & 0.659 $\pm$ 0.031 & 0.649 $\pm$ 0.043 & 0.835 $\pm$ 0.024 & 0.659 $\pm$ 0.027 \\
         (b) MOTCat w/o OT & ~ &$\checkmark$& 0.663 $\pm$ 0.025 & 0.666 $\pm$ 0.017 & 0.663 $\pm$ 0.031 & 0.845 $\pm$ 0.023 & \textbf{0.673 $\pm$ 0.036} \\
         \textbf{(c) MOTCat }& $\checkmark$ & $\checkmark$ & \textbf{0.683 $\pm$ 0.026} & \textbf{0.673 $\pm$ 0.006} & \textbf{0.675 $\pm$ 0.040} & \textbf{0.849 $\pm$ 0.028} & 0.670 $\pm$ 0.038 \\
         \bottomrule
    \end{tabular}
    \label{tab:abla}
\end{table*}

\begin{figure*}[t]
    \centering
    \includegraphics[scale=0.53]{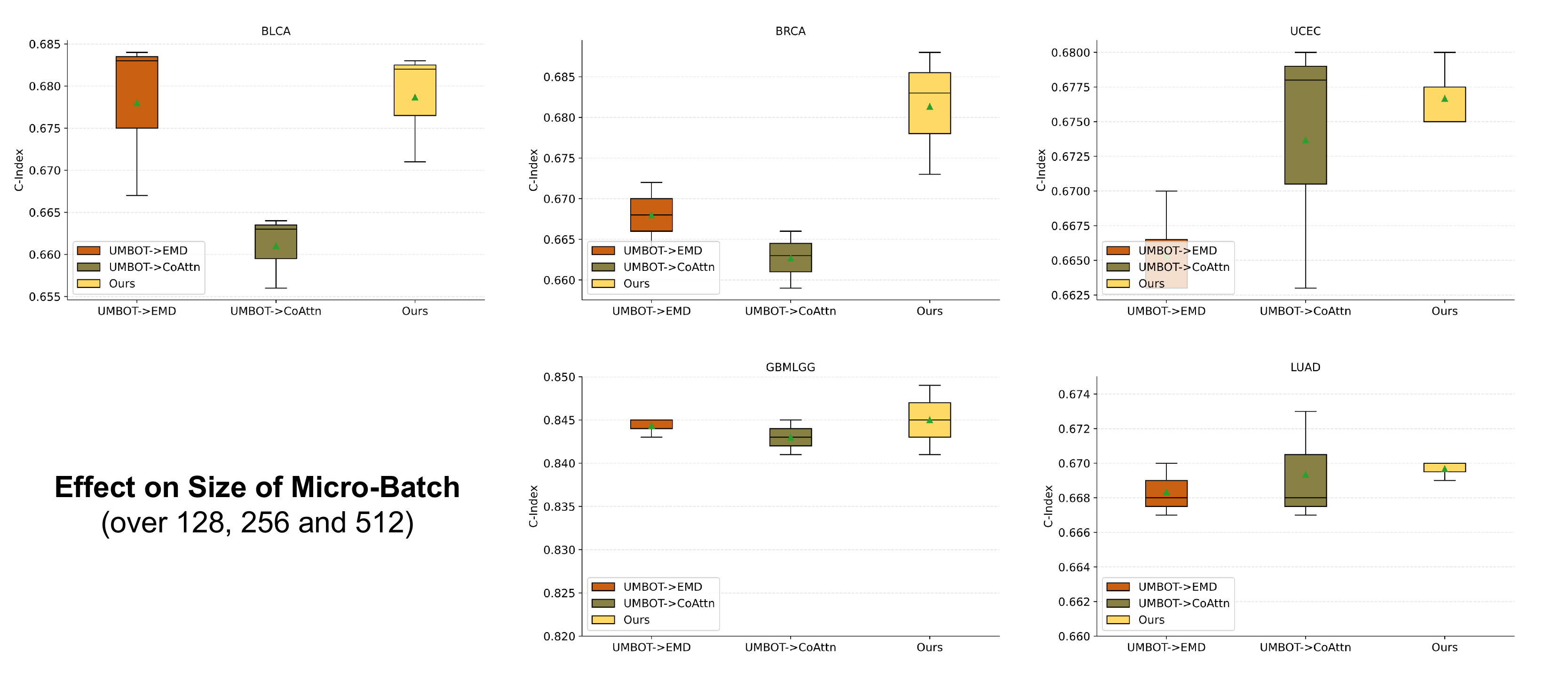}
    \caption{Boxplots of C-Index over various sizes of Micro-batch for our method and two variants on five cancer datasets, in which the green triangle means the averaged results of all sizes and the horizontal line in the box represents the median result.}
    \label{fig:box_bs}
\end{figure*}



\begin{figure*}[h]
    \centering
    \includegraphics[scale=0.5]{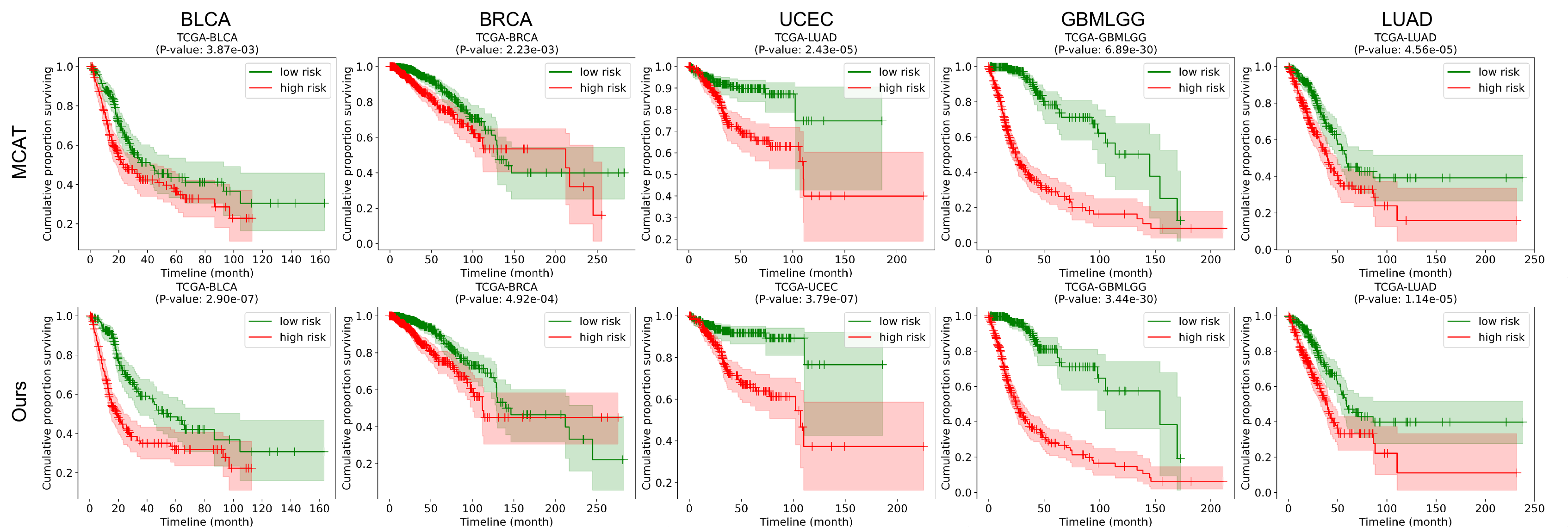}
    \caption{Kaplan-Meier Analysis on five cancer datasets, where patient stratifications of low risk (green) and high risk (red) are presented. Shaded areas refer to the confidence intervals. $\text{P-value} <0.05$ means significant statistical difference in two groups, and lower P-value is better. (Zoom in for better viewing.)}
    \label{fig:km}
\end{figure*}
We first investigate the effectiveness of OT-based co-attention component (denoted by OT) and the micro-batch strategy (denoted by MB). Furthermore, the effect on the size of micro-batch and the actual computational speed of our method are explored.

\noindent
\textbf{Component validation.} As shown in Tab.~\ref{tab:abla}, we observe that multimodal fusion can benefit from micro-batch strategy by comparing (a) and (b), and OT-based co-attention further improves the performance by approximating the original results with the averaged results over micro-batches via UMBOT. As a result, our method achieves the best overall performance. These improvements both demonstrate the effectiveness of each component for the proposed method. 

Additionally, recent work~\cite{zhang2022dtfd} has claimed that MIL on WSIs can benefit from learning over sub-bags in a WSI, which is also validated by performance increases in variant (b). The reason why it can profit from micro-batch might be that it increases the number of bags, so that more features can be used for survival analysis.

\noindent
\textbf{Size of Micro-batch.} To show the robustness of approximation to the size of Micro-batch strategy, we compare several variants of the proposed method over the varying sizes of micro-batch, as shown in the quantitative analysis of Fig.~\ref{fig:box_bs}. The qualitative analysis of the effect on size of Micro-Batch can be found in the supplementary materials, which also achieves the consistent conclusion.

For the quantitative analysis in Fig.~\ref{fig:box_bs}, two variants of the proposed method are compared: 1) \textbf{UMBOT $\rightarrow$ EMD}: To demonstrate the robustness of different OT variants, we replace UMBOT used in our method with the original OT (i.e. EMD) for comparison, which is equivalent to the Eq. (4) done over Micro-batch. 2) \textbf{UMBOT $\rightarrow$ CoAttn}: To compare the robustness of the proposed OT-based co-attention and the original co-attention, we replace UMBOT with the co-attention used in MCAT~\cite{chen2021multimodal} and train the model over Micro-batch for comparison.

From the results of Fig.~\ref{fig:box_bs}, we can see that 1) our method achieves the best averaged performance of various sizes compared with the two variants, especially on BRCA with the most samples of 956. 2) Further, on UCEC and LUAD datasets, the proposed method gets the most robust results. 3) Although the second variant UMBOT $\rightarrow$ CoAttn shows slightly better robustness on BLCA and BRCA, the performance of our method surpasses it by a large margin. In a nutshell, the proposed method achieves a better trade-off between performance and robustness, compared with different OT variants or the co-attention of MCAT.

\noindent
\textbf{Computational Speed.} We measure the actual averaged computational speed for 10 WSIs (about 150k patches in total) on one GPU of NVIDIA GeForce RTX 3090. The proposed method is compared with a variant 'org-OT' that replaces UMBOT by the original OT. However, the original OT with the high complexity takes extreme long time for one WSI (about 20k patches), and thus we fail to measure the actual computational time for the original OT in a short time. Here we present the speed of our method, which suggests our method makes it practicable to apply OT to histology and genomics. Specifically, the training speed of the proposed method is 6540 p/s (p/s refers to the number of processing patches per second) and the inference speed is 11885 p/s.

\subsection{Statistical Analysis}
To show a statistical difference in patient stratification performance, we visualize the Kaplan-Meier survival curves for different methods, in which patients are separated into two groups of low-risk and high-risk based on predicted risk scores, and then the statistics on ground-truth survival time are presented for each group in Fig.~\ref{fig:km}. The Logrank test is conducted to measure the statistically significant difference in two groups of patients, where a lower P-value indicates better performance of patient stratification. From the results of Fig.~\ref{fig:km}, intuitively our method separates patients of low and high risk more clearly on all datasets. In the Logrank test, our method achieves a lower P-value on all datasets in comparison to MCAT, especially on BLCA, BRCA and UCEC with a large margin of magnitude.

\section{Conclusion}
In this paper, we present a novel Multimodal Optimal Transport-based Co-Attention Transformer with global structure consistency to tackle two important issues in multimodal learning of histology and genomics for survival prediction. First, we utilize a new OT-based Co-Attention to match pairwise instances between histology and genomics for selecting informative instances strongly related to tumor microenvironment, which provides a way to effectively represent gigapixel WSIs. Second, optimal transport offers a global awareness for modeling potential structure within modality, i.e. pathological interactions and genomic co-expression. Furthermore, to enable applying OT-based Co-Attention in practice, we propose a robust and efficient implementation over micro-batch of WSI bags, in which the approximation over micro-batch might provide a solution to update the parameters of extractor for patch images in an end-to-end way instead of extracting feature embeddings offline, which will be further explored in the future.

\section*{Acknowledgement}
This work was supported by National Natural Science Foundation of China (No. 62202403), Shenzhen Science and Technology Innovation Committee (Project No. SGDX20210823103201011) and Hong Kong Innovation and Technology Fund (No. PRP/034/22FX).
{\small
\bibliographystyle{ieee_fullname}
\bibliography{egbib}

\begin{thebibliography}{10}\itemsep=-1pt

\bibitem{abduljabbar2020geospatial}
Khalid AbdulJabbar, Shan E~Ahmed Raza, Rachel Rosenthal, Mariam Jamal-Hanjani,
  Selvaraju Veeriah, Ayse Akarca, Tom Lund, David~A Moore, Roberto Salgado,
  Maise Al~Bakir, et~al.
\newblock Geospatial immune variability illuminates differential evolution of
  lung adenocarcinoma.
\newblock {\em Nature medicine}, 26(7):1054--1062, 2020.

\bibitem{campanella2019clinical}
Gabriele Campanella, Matthew~G Hanna, Luke Geneslaw, Allen Miraflor, Vitor
  Werneck Krauss~Silva, Klaus~J Busam, Edi Brogi, Victor~E Reuter, David~S
  Klimstra, and Thomas~J Fuchs.
\newblock Clinical-grade computational pathology using weakly supervised deep
  learning on whole slide images.
\newblock {\em Nature medicine}, 25(8):1301--1309, 2019.

\bibitem{caootkge2022otkge}
Zongsheng Cao, Qianqian Xu, Zhiyong Yang, Yuan He, Xiaochun Cao, and Qingming
  Huang.
\newblock Otkge: Multi-modal knowledge graph embeddings via optimal transport.
\newblock In {\em Advances in Neural Information Processing Systems}, 2022.

\bibitem{richard2022tmi}
Richard~J. Chen, Ming~Y. Lu, Jingwen Wang, Drew F.~K. Williamson, Scott~J.
  Rodig, Neal~I. Lindeman, and Faisal Mahmood.
\newblock Pathomic fusion: An integrated framework for fusing histopathology
  and genomic features for cancer diagnosis and prognosis.
\newblock {\em IEEE Transactions on Medical Imaging}, 41(4):757--770, 2022.

\bibitem{chen2021multimodal}
Richard~J Chen, Ming~Y Lu, Wei-Hung Weng, Tiffany~Y Chen, Drew~FK Williamson,
  Trevor Manz, Maha Shady, and Faisal Mahmood.
\newblock Multimodal co-attention transformer for survival prediction in
  gigapixel whole slide images.
\newblock In {\em Proceedings of the IEEE/CVF International Conference on
  Computer Vision}, pages 4015--4025, 2021.

\bibitem{chen2022pan}
Richard~J Chen, Ming~Y Lu, Drew~FK Williamson, Tiffany~Y Chen, Jana Lipkova,
  Zahra Noor, Muhammad Shaban, Maha Shady, Mane Williams, Bumjin Joo, et~al.
\newblock Pan-cancer integrative histology-genomic analysis via multimodal deep
  learning.
\newblock {\em Cancer Cell}, 40(8):865--878, 2022.

\bibitem{chikontwe2020multiple}
Philip Chikontwe, Meejeong Kim, Soo~Jeong Nam, Heounjeong Go, and Sang~Hyun
  Park.
\newblock Multiple instance learning with center embeddings for histopathology
  classification.
\newblock In {\em Medical Image Computing and Computer Assisted
  Intervention--MICCAI 2020: 23rd International Conference, Lima, Peru, October
  4--8, 2020, Proceedings, Part V 23}, pages 519--528. Springer, 2020.

\bibitem{chizat2018scaling}
Lenaic Chizat, Gabriel Peyr{\'e}, Bernhard Schmitzer, and Fran{\c{c}}ois-Xavier
  Vialard.
\newblock Scaling algorithms for unbalanced optimal transport problems.
\newblock {\em Mathematics of Computation}, 87(314):2563--2609, 2018.

\bibitem{david1972regression}
Cox~R David et~al.
\newblock Regression models and life tables (with discussion).
\newblock {\em Journal of the Royal Statistical Society}, 34(2):187--220, 1972.

\bibitem{deng2009imagenet}
Jia Deng, Wei Dong, Richard Socher, Li-Jia Li, Kai Li, and Li Fei-Fei.
\newblock Imagenet: A large-scale hierarchical image database.
\newblock In {\em 2009 IEEE conference on computer vision and pattern
  recognition}, pages 248--255. Ieee, 2009.

\bibitem{duan2022multi}
Jiali Duan, Liqun Chen, Son Tran, Jinyu Yang, Yi Xu, Belinda Zeng, and Trishul
  Chilimbi.
\newblock Multi-modal alignment using representation codebook.
\newblock In {\em Proceedings of the IEEE/CVF Conference on Computer Vision and
  Pattern Recognition}, pages 15651--15660, 2022.

\bibitem{faraggi1995neural}
David Faraggi and Richard Simon.
\newblock A neural network model for survival data.
\newblock {\em Statistics in medicine}, 14(1):73--82, 1995.

\bibitem{fatras2021unbalanced}
Kilian Fatras, Thibault S{\'e}journ{\'e}, R{\'e}mi Flamary, and Nicolas Courty.
\newblock Unbalanced minibatch optimal transport; applications to domain
  adaptation.
\newblock In {\em International Conference on Machine Learning}, pages
  3186--3197. PMLR, 2021.

\bibitem{frogner2015learning}
Charlie Frogner, Chiyuan Zhang, Hossein Mobahi, Mauricio Araya, and Tomaso~A
  Poggio.
\newblock Learning with a wasserstein loss.
\newblock {\em Advances in neural information processing systems}, 28, 2015.

\bibitem{gadzicki2020early}
Konrad Gadzicki, Razieh Khamsehashari, and Christoph Zetzsche.
\newblock Early vs late fusion in multimodal convolutional neural networks.
\newblock In {\em 2020 IEEE 23rd international conference on information fusion
  (FUSION)}, pages 1--6. IEEE, 2020.

\bibitem{hashimoto2020multi}
Noriaki Hashimoto, Daisuke Fukushima, Ryoichi Koga, Yusuke Takagi, Kaho Ko, Kei
  Kohno, Masato Nakaguro, Shigeo Nakamura, Hidekata Hontani, and Ichiro
  Takeuchi.
\newblock Multi-scale domain-adversarial multiple-instance cnn for cancer
  subtype classification with unannotated histopathological images.
\newblock In {\em Proceedings of the IEEE/CVF conference on computer vision and
  pattern recognition}, pages 3852--3861, 2020.

\bibitem{he2016identity}
Kaiming He, Xiangyu Zhang, Shaoqing Ren, and Jian Sun.
\newblock Identity mappings in deep residual networks.
\newblock In {\em Computer Vision--ECCV 2016: 14th European Conference,
  Amsterdam, The Netherlands, October 11--14, 2016, Proceedings, Part IV 14},
  pages 630--645. Springer, 2016.

\bibitem{huang2020fusion}
Shih-Cheng Huang, Anuj Pareek, Saeed Seyyedi, Imon Banerjee, and Matthew~P
  Lungren.
\newblock Fusion of medical imaging and electronic health records using deep
  learning: a systematic review and implementation guidelines.
\newblock {\em NPJ digital medicine}, 3(1):136, 2020.

\bibitem{ilse2018attention}
Maximilian Ilse, Jakub Tomczak, and Max Welling.
\newblock Attention-based deep multiple instance learning.
\newblock In {\em International conference on machine learning}, pages
  2127--2136. PMLR, 2018.

\bibitem{joo2021multimodal}
Sunghoon Joo, Eun~Sook Ko, Soonhwan Kwon, Eunjoo Jeon, Hyungsik Jung, Ji-Yeon
  Kim, Myung~Jin Chung, and Young-Hyuck Im.
\newblock Multimodal deep learning models for the prediction of pathologic
  response to neoadjuvant chemotherapy in breast cancer.
\newblock {\em Scientific reports}, 11(1):18800, 2021.

\bibitem{kantorovich2006translocation}
Leonid~V Kantorovich.
\newblock On the translocation of masses.
\newblock {\em Journal of mathematical sciences}, 133(4):1381--1382, 2006.

\bibitem{klambauer2017self}
G{\"u}nter Klambauer, Thomas Unterthiner, Andreas Mayr, and Sepp Hochreiter.
\newblock Self-normalizing neural networks.
\newblock {\em Advances in neural information processing systems}, 30, 2017.

\bibitem{2011Survival}
D.~G. Kleinbaum and M. Klein.
\newblock {\em Survival analysis : a self-learning text / 2nd ed}.
\newblock Survival analysis : a self-learning text / 2nd ed, 2011.

\bibitem{kumar2019co}
Ashnil Kumar, Michael Fulham, Dagan Feng, and Jinman Kim.
\newblock Co-learning feature fusion maps from pet-ct images of lung cancer.
\newblock {\em IEEE Transactions on Medical Imaging}, 39(1):204--217, 2019.

\bibitem{kuroda2021tumor}
Hajime Kuroda, Tsengelmaa Jamiyan, Rin Yamaguchi, Akinari Kakumoto, Akihito
  Abe, Oi Harada, and Atsuko Masunaga.
\newblock Tumor-infiltrating b cells and t cells correlate with postoperative
  prognosis in triple-negative carcinoma of the breast.
\newblock {\em BMC cancer}, 21(1):1--10, 2021.

\bibitem{li2021dual}
Bin Li, Yin Li, and Kevin~W Eliceiri.
\newblock Dual-stream multiple instance learning network for whole slide image
  classification with self-supervised contrastive learning.
\newblock In {\em Proceedings of the IEEE/CVF conference on computer vision and
  pattern recognition}, pages 14318--14328, 2021.

\bibitem{li2022hfbsurv}
Ruiqing Li, Xingqi Wu, Ao Li, and Minghui Wang.
\newblock Hfbsurv: hierarchical multimodal fusion with factorized bilinear
  models for cancer survival prediction.
\newblock {\em Bioinformatics}, 38(9):2587--2594, 2022.

\bibitem{liberzon2015molecular}
Arthur Liberzon, Chet Birger, Helga Thorvaldsd{\'o}ttir, Mahmoud Ghandi, Jill~P
  Mesirov, and Pablo Tamayo.
\newblock The molecular signatures database hallmark gene set collection.
\newblock {\em Cell systems}, 1(6):417--425, 2015.

\bibitem{lin2019ghrelin}
Tsung-Chieh Lin, Yuan-Ming Yeh, Wen-Lang Fan, Yu-Chan Chang, Wei-Ming Lin,
  Tse-Yen Yang, and Michael Hsiao.
\newblock Ghrelin upregulates oncogenic aurora a to promote renal cell
  carcinoma invasion.
\newblock {\em Cancers}, 11(3):303, 2019.

\bibitem{lipkova2022artificial}
Jana Lipkova, Richard~J Chen, Bowen Chen, Ming~Y Lu, Matteo Barbieri, Daniel
  Shao, Anurag~J Vaidya, Chengkuan Chen, Luoting Zhuang, Drew~FK Williamson,
  et~al.
\newblock Artificial intelligence for multimodal data integration in oncology.
\newblock {\em Cancer Cell}, 40(10):1095--1110, 2022.

\bibitem{liu2021tumor}
Dan Liu, Xue Yang, and Xiongzhi Wu.
\newblock Tumor immune microenvironment characterization identifies prognosis
  and immunotherapy-related gene signatures in melanoma.
\newblock {\em Frontiers in immunology}, 12:663495, 2021.

\bibitem{lu2021data}
Ming~Y Lu, Drew~FK Williamson, Tiffany~Y Chen, Richard~J Chen, Matteo Barbieri,
  and Faisal Mahmood.
\newblock Data-efficient and weakly supervised computational pathology on
  whole-slide images.
\newblock {\em Nature biomedical engineering}, 5(6):555--570, 2021.

\bibitem{oya2020tumor}
Yukiko Oya, Yoku Hayakawa, and Kazuhiko Koike.
\newblock Tumor microenvironment in gastric cancers.
\newblock {\em Cancer science}, 111(8):2696--2707, 2020.

\bibitem{qiu2023deep}
Lin Qiu, Aminollah Khormali, and Kai Liu.
\newblock Deep biological pathway informed pathology-genomic multimodal
  survival prediction.
\newblock {\em arXiv preprint arXiv:2301.02383}, 2023.

\bibitem{ramachandram2017deep}
Dhanesh Ramachandram and Graham~W Taylor.
\newblock Deep multimodal learning: A survey on recent advances and trends.
\newblock {\em IEEE signal processing magazine}, 34(6):96--108, 2017.

\bibitem{shao2021transmil}
Zhuchen Shao, Hao Bian, Yang Chen, Yifeng Wang, Jian Zhang, Xiangyang Ji,
  et~al.
\newblock Transmil: Transformer based correlated multiple instance learning for
  whole slide image classification.
\newblock {\em Advances in neural information processing systems},
  34:2136--2147, 2021.

\bibitem{sharma2021cluster}
Yash Sharma, Aman Shrivastava, Lubaina Ehsan, Christopher~A Moskaluk, Sana
  Syed, and Donald Brown.
\newblock Cluster-to-conquer: A framework for end-to-end multi-instance
  learning for whole slide image classification.
\newblock In {\em Medical Imaging with Deep Learning}, pages 682--698. PMLR,
  2021.

\bibitem{wang2019mmp}
Qi-Min Wang, LI Lv, Ying Tang, LI Zhang, and Li-Fen Wang.
\newblock Mmp-1 is overexpressed in triple-negative breast cancer tissues and
  the knockdown of mmp-1 expression inhibits tumor cell malignant behaviors in
  vitro.
\newblock {\em Oncology letters}, 17(2):1732--1740, 2019.

\bibitem{wang2020preoperative}
Siwen Wang, Caizhen Feng, Di Dong, Hailin Li, Jing Zhou, Yingjiang Ye, Zaiyi
  Liu, Jie Tian, and Yi Wang.
\newblock Preoperative computed tomography-guided disease-free survival
  prediction in gastric cancer: a multicenter radiomics study.
\newblock {\em Medical Physics}, 47(10):4862--4871, 2020.

\bibitem{wang2019weakly}
Xi Wang, Hao Chen, Caixia Gan, Huangjing Lin, Qi Dou, Efstratios Tsougenis,
  Qitao Huang, Muyan Cai, and Pheng-Ann Heng.
\newblock Weakly supervised deep learning for whole slide lung cancer image
  analysis.
\newblock {\em IEEE transactions on cybernetics}, 50(9):3950--3962, 2019.

\bibitem{wang2021wasserstein}
Yun Wang, Tong Zhang, Xueya Zhang, Zhen Cui, Yuge Huang, Pengcheng Shen,
  Shaoxin Li, and Jian Yang.
\newblock Wasserstein coupled graph learning for cross-modal retrieval.
\newblock In {\em 2021 IEEE/CVF International Conference on Computer Vision
  (ICCV)}, pages 1793--1802. IEEE, 2021.

\bibitem{wang2022efficient}
Zongjie Wang, Sharif Ahmed, Mahmoud Labib, Hansen Wang, Xiyue Hu, Jiarun Wei,
  Yuxi Yao, Jason Moffat, Edward~H Sargent, and Shana~O Kelley.
\newblock Efficient recovery of potent tumour-infiltrating lymphocytes through
  quantitative immunomagnetic cell sorting.
\newblock {\em Nature Biomedical Engineering}, 6(2):108--117, 2022.

\bibitem{wang2021gpdbn}
Zhiqin Wang, Ruiqing Li, Minghui Wang, and Ao Li.
\newblock Gpdbn: deep bilinear network integrating both genomic data and
  pathological images for breast cancer prognosis prediction.
\newblock {\em Bioinformatics}, 37(18):2963--2970, 2021.

\bibitem{xu2019camel}
Gang Xu, Zhigang Song, Zhuo Sun, Calvin Ku, Zhe Yang, Cancheng Liu, Shuhao
  Wang, Jianpeng Ma, and Wei Xu.
\newblock Camel: A weakly supervised learning framework for histopathology
  image segmentation.
\newblock In {\em Proceedings of the IEEE/CVF International Conference on
  computer vision}, pages 10682--10691, 2019.

\bibitem{yao2020whole}
Jiawen Yao, Xinliang Zhu, Jitendra Jonnagaddala, Nicholas Hawkins, and Junzhou
  Huang.
\newblock Whole slide images based cancer survival prediction using attention
  guided deep multiple instance learning networks.
\newblock {\em Medical Image Analysis}, 65:101789, 2020.

\bibitem{zadeh2020bias}
Shekoufeh~Gorgi Zadeh and Matthias Schmid.
\newblock Bias in cross-entropy-based training of deep survival networks.
\newblock {\em IEEE transactions on pattern analysis and machine intelligence},
  43(9):3126--3137, 2020.

\bibitem{zeng2021exploration}
Yangyang Zeng, Yulan Zeng, Hang Yin, Fengxia Chen, Qingqing Wang, Xiaoyan Yu,
  and Yunfeng Zhou.
\newblock Exploration of the immune cell infiltration-related gene signature in
  the prognosis of melanoma.
\newblock {\em Aging (albany NY)}, 13(3):3459, 2021.

\bibitem{zhang2022dtfd}
Hongrun Zhang, Yanda Meng, Yitian Zhao, Yihong Qiao, Xiaoyun Yang, Sarah~E
  Coupland, and Yalin Zheng.
\newblock Dtfd-mil: Double-tier feature distillation multiple instance learning
  for histopathology whole slide image classification.
\newblock In {\em Proceedings of the IEEE/CVF Conference on Computer Vision and
  Pattern Recognition}, pages 18802--18812, 2022.

\bibitem{zhang2014normalized}
Jie Zhang and Kun Huang.
\newblock Normalized imqcm: An algorithm for detecting weak quasi-cliques in
  weighted graph with applications in gene co-expression module discovery in
  cancers.
\newblock {\em Cancer informatics}, 13:CIN--S14021, 2014.

\bibitem{zheng2022multi}
Hanci Zheng, Zongying Lin, Qizheng Zhou, Xingchen Peng, Jianghong Xiao, Chen
  Zu, Zhengyang Jiao, and Yan Wang.
\newblock Multi-transsp: Multimodal transformer for survival prediction of
  nasopharyngeal carcinoma patients.
\newblock In {\em Medical Image Computing and Computer Assisted
  Intervention--MICCAI 2022: 25th International Conference, Singapore,
  September 18--22, 2022, Proceedings, Part VII}, pages 234--243. Springer,
  2022.

\bibitem{zhou2021computational}
Meng Zhou, Zicheng Zhang, Siqi Bao, Ping Hou, Congcong Yan, Jianzhong Su, and
  Jie Sun.
\newblock Computational recognition of lncrna signature of tumor-infiltrating b
  lymphocytes with potential implications in prognosis and immunotherapy of
  bladder cancer.
\newblock {\em Briefings in Bioinformatics}, 22(3):bbaa047, 2021.

\bibitem{zhu2022multimodal}
Qi Zhu, Heyang Wang, Bingliang Xu, Zhiqiang Zhang, Wei Shao, and Daoqiang
  Zhang.
\newblock Multimodal triplet attention network for brain disease diagnosis.
\newblock {\em IEEE Transactions on Medical Imaging}, 41(12):3884--3894, 2022.

\bibitem{zhu2016deep}
Xinliang Zhu, Jiawen Yao, and Junzhou Huang.
\newblock Deep convolutional neural network for survival analysis with
  pathological images.
\newblock In {\em 2016 IEEE International Conference on Bioinformatics and
  Biomedicine (BIBM)}, pages 544--547. IEEE, 2016.

\bibitem{zhu2017wsisa}
Xinliang Zhu, Jiawen Yao, Feiyun Zhu, and Junzhou Huang.
\newblock Wsisa: Making survival prediction from whole slide histopathological
  images.
\newblock In {\em Proceedings of the IEEE conference on computer vision and
  pattern recognition}, pages 7234--7242, 2017.

\bibitem{zuo2022identify}
Yingli Zuo, Yawen Wu, Zixiao Lu, Qi Zhu, Kun Huang, Daoqiang Zhang, and Wei
  Shao.
\newblock Identify consistent imaging genomic biomarkers for characterizing the
  survival-associated interactions between tumor-infiltrating lymphocytes and
  tumors.
\newblock In {\em Medical Image Computing and Computer Assisted
  Intervention--MICCAI 2022: 25th International Conference, Singapore,
  September 18--22, 2022, Proceedings, Part II}, pages 222--231. Springer,
  2022.

\end{thebibliography}
}

\appendix
\section{Outline}
The supplementary materials for this paper are organized as follows:
\begin{enumerate}

\item We demonstrate the qualitative visualization for the effect on the size of Micro-Batch Strategy.
\item We provide visualization cases of Co-Attention.
\end{enumerate}

\section{Effect on Size of Micro-Batch}
\begin{figure*}
    \centering
    \includegraphics[scale=0.5]{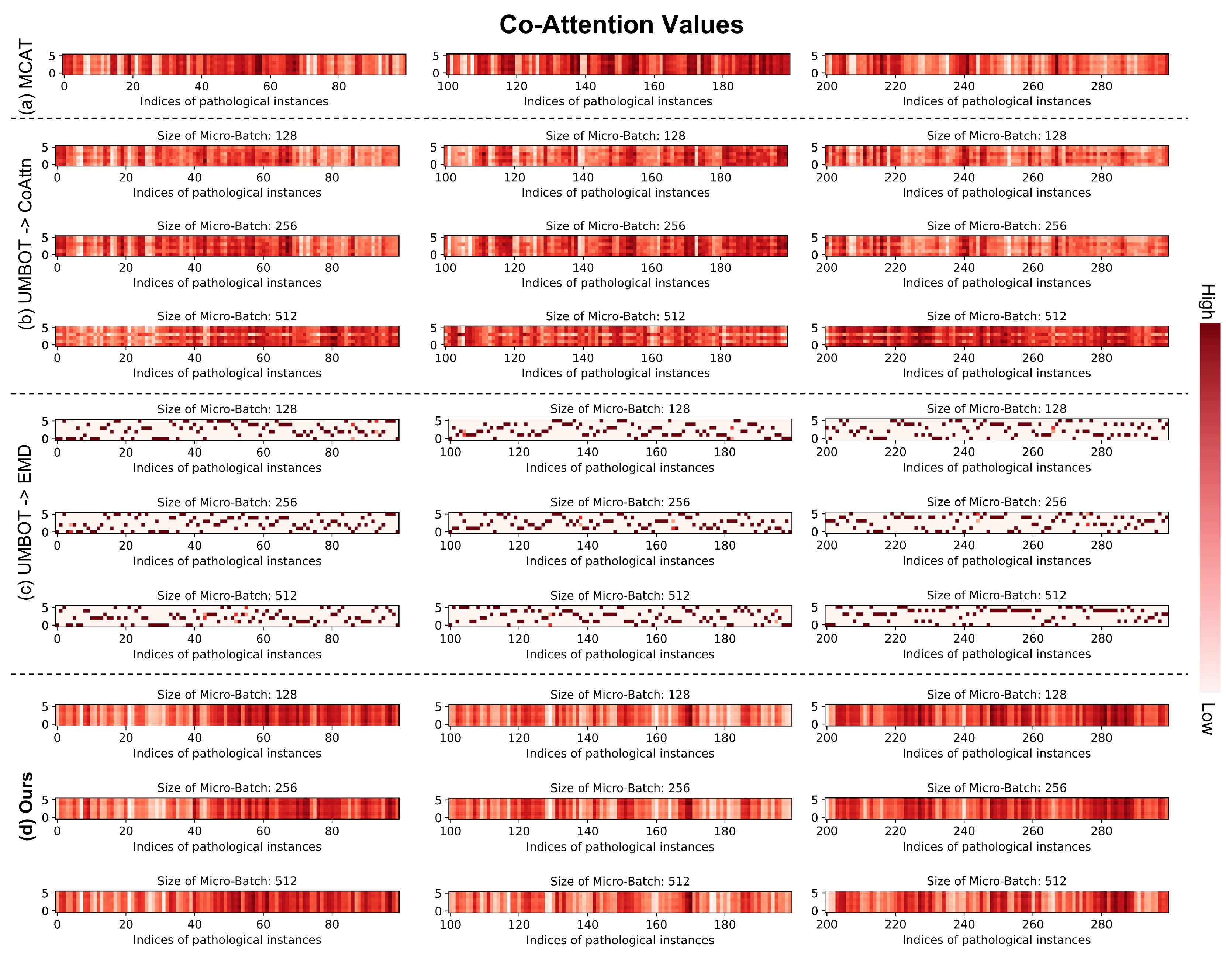}
    \caption{Co-Attention values between the first 300 instances of histology and all instances of genomics for the case \textit{TCGA-06-0210} of GBMLGG: (a) MCAT, (b) UMBOT $\rightarrow$ CoAttn, (c) UMBOT $\rightarrow$ EMD and (d) our method, in which the size of Micro-Batch ranges from 128 to 512.}
    \label{fig:coattn}
\end{figure*}

In this section, we compare several variants of the proposed method to show the robustness to the size of Micro-batch strategy used in histology data, in which the results over the size of 128, 256 and 512 are presented in Fig.~\ref{fig:coattn}.

There are two variants of the proposed method mentioned in the quantitative analysis of Section 4.3 and MCAT to be compared for the qualitative analysis. As shown in Fig.~\ref{fig:coattn}, we visualize the co-attention values between the first 300 instances of a WSI and all genomic instances ($M_g=6$) from the same patient.

We observe that 1) our method demonstrates the best consistency of activation among different sizes of Micro-Batch, while the variant (c) UMBOT $\rightarrow$ EMD shows considerably poor consistency, which validates that UMBOT-based co-attention used in our method is more robust to the size of Micro-Batch than the original OT (i.e. EMD). 2) When we replace UMBOT with the co-attention used in MCAT and train it over Micro-Batch, we found that the activation pattern of size 512 is significantly different from that of sizes 128 and 256, as shown in the variant (b) UMBOT $\rightarrow$ CoAttn. 3) Furthermore, results of all sizes in variant (b) are apparently distinguished from the co-attention of MCAT directly computed over all instances, further indicating the poor robustness of the original co-attention.
\section{Visualization of Co-Attention}
\begin{figure*}
    \centering
    \includegraphics[scale=0.65]{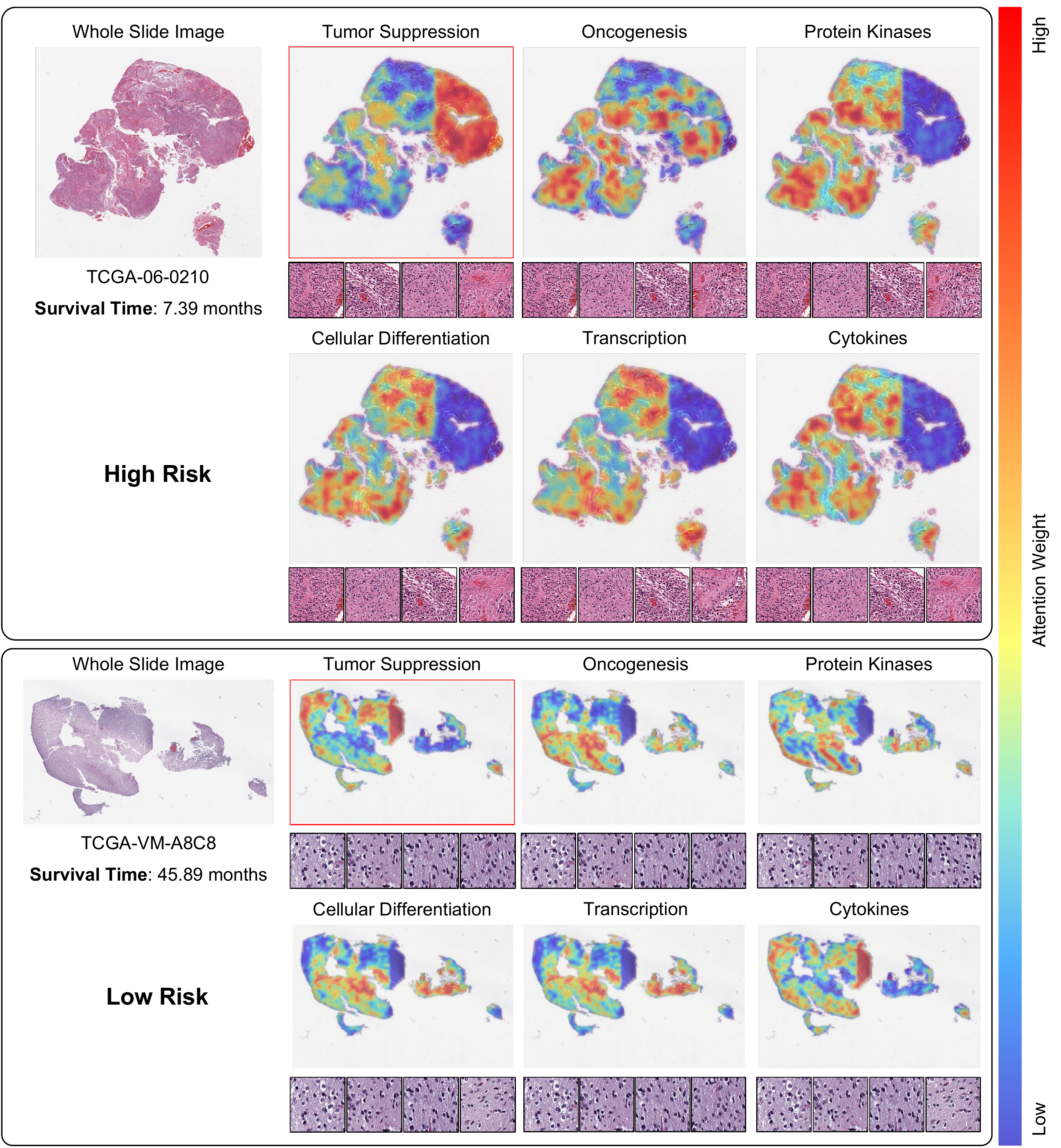}
    \caption{Co-Attention visualization of \textbf{our method} for high and low risk cases in GBMLGG, with corresponding top-4 highest attention patches for each genomic instance of unique functional category.}
    \label{fig:vis_motcat}
\end{figure*}

\begin{figure*}
    \centering
    \includegraphics[scale=0.65]{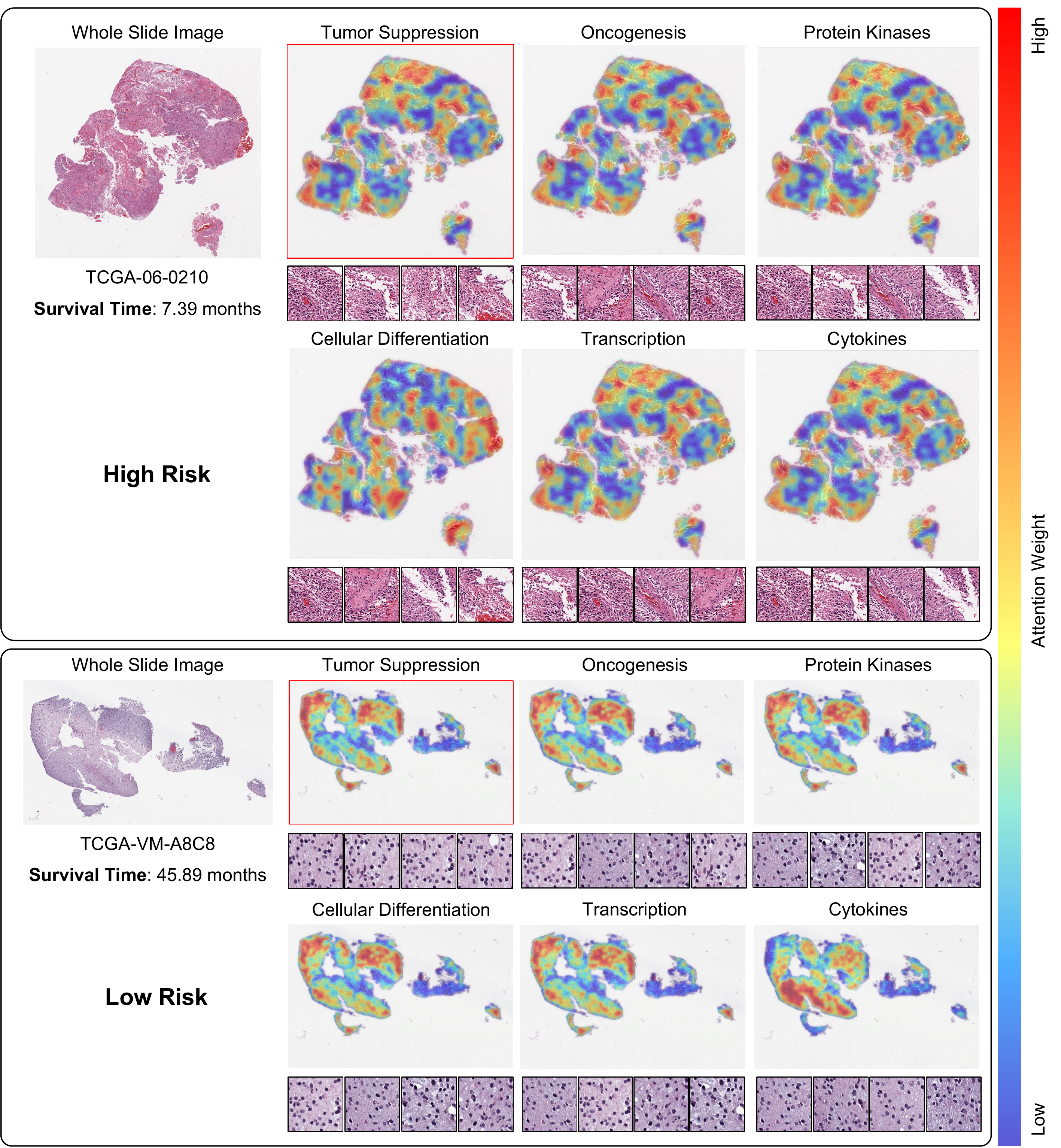}
    \caption{Co-Attention visualization of \textbf{MCAT}~\cite{chen2021multimodal} for high and low risk cases in GBMLGG, with corresponding top-4 highest attention patches for each genomic instance of unique functional category.}
    \label{fig:vis_mcat}
\end{figure*}

To show the interpretability, we visualize the co-attention values of all instances in each WSI for high and low cases, as shown in Fig.~\ref{fig:vis_motcat} of our method and Fig.~\ref{fig:vis_mcat} of MCAT~\cite{chen2021multimodal}. In order to present a more obvious difference in co-attention values among various genomic instances, we consider the instance of Tumor Suppression (marked in red) as the reference and show the differences of other genomic instances from it, since there is only a slight difference in the values of co-attention.

By comparing MCAT with the proposed method, we found that the OT-based co-attention is concerned about different areas of the WSI for different genomic functional instances, while the dense co-attention used in MCAT focuses on the similar regions of histology for different functional instances of genes. As a result, the better performance of our method may benefit from these different concerns.

\end{document}